\documentclass[sigconf,nonacm]{acmart}

\AtBeginDocument{%
  \providecommand\BibTeX{{%
    \normalfont B\kern-0.5em{\scshape i\kern-0.25em b}\kern-0.8em\TeX}}}

\setcopyright{none}
\copyrightyear{}
\acmYear{}
\acmDOI{}

\acmConference[]{}{}{}
\acmBooktitle{}
\acmPrice{}
\acmISBN{}

\acmSubmissionID{514}

\usepackage{subfigure}
\usepackage{multirow}
\usepackage{cleveref}
\usepackage{math}
\usepackage{epsfig}
\usepackage{graphicx}
\usepackage{amsmath}
\usepackage{amssymb}
\usepackage{subfigure}
\usepackage{multirow}
\usepackage{enumitem}
\usepackage{adjustbox}
\usepackage{amsmath}
\usepackage{amssymb}
\usepackage{bm}
\usepackage{xcolor}
\newcommand{\red}[1]{#1}
\newcommand{\blue}[1]{#1}

\usepackage{xcolor}

\crefdefaultlabelformat{#2\!#1#3}
\crefname{figure}{Fig.}{Figs.}
\Crefname{figure}{Fig.}{Figs.}
\crefname{equation}{Eq.}{Eqs.}
\Crefname{equation}{Eq.}{Eqs.}
\crefname{table}{Tab.}{Tabs.}
\Crefname{table}{Tab.}{Tabs.}
\crefname{section}{Sec.}{Secs.}
\Crefname{section}{Sec.}{Secs.}

\usepackage{pifont}
\usepackage[group-separator={,}]{siunitx}

\begin{document}

\title{Learning Task-oriented Disentangled Representations for Unsupervised Domain Adaptation}


\author{Pingyang Dai}
\affiliation{\institution{Xiamen University}}
\email{pydai@xmu.edu.cn}

\author{Peixian Chen}
\affiliation{\institution{Xiamen University}}
\email{pxchen@stu.xmu.edu.cn}

\author{Qiong Wu}
\affiliation{\institution{Xiamen University}}
\email{qiong@stu.xmu.edu.cn}

\author{Xiaopeng Hong}
\affiliation{\institution{Xi'an Jiaotong University}}
\email{hongxiaopeng@mail.xjtu.edu.cn}

\author{Qixiang Ye}
\affiliation{\institution{University of Chinese Academy of Sciences, China}}
\email{qxye@ucas.ac.cn}

\author{Qi Tian}
\affiliation{\institution{Huawei Cloud \& AI}}
\email{tian.qi1@huawei.com}

\author{Rongrong Ji}
\affiliation{\institution{Xiamen University, China}}
\email{rrji@xmu.edu.cn}


	\begin{abstract}
		Unsupervised domain adaptation (UDA) aims to address the domain-shift problem between a labeled source domain and an unlabeled target domain.
		%
		\blue{Many efforts have been made to eliminate the mismatch between the distributions of training and testing data by learning \red{domain-invariant} representations. However, the learned representations are usually not task-oriented, i.e., being class-discriminative and domain-transferable simultaneously. This drawback limits the flexibility of UDA in complicated open-set tasks where no labels are shared between domains.}
		In this paper, we break the concept of task-orientation into task-relevance and task-irrelevance, and propose a dynamic task-oriented disentangling network (DTDN) to learn disentangled representations in an end-to-end fashion for \red{UDA}.
		The dynamic disentangling network effectively disentangles data representations into two components: the task-relevant ones embedding critical information associated with the task across domains, and the task-irrelevant ones with the remaining non-transferable or disturbing information.
		These two components are regularized by a group of task-specific objective functions across domains. Such regularization explicitly encourages disentangling and avoids the use of generative models or decoders.
		\blue{Experiments in complicated, open-set scenarios (retrieval tasks) and empirical benchmarks (classification tasks) demonstrate that the proposed method captures rich disentangled information and achieves superior performance.}
	\end{abstract}

\maketitle

	\section{Introduction}
	Despite the exciting prominence in various computer vision and machine learning problems, the learning of deep representations heavily relies on the availability of large-scale labeled data in a specific domain.
	Such a setting is prohibitive for many real-world applications, considering that large-scale annotations are costly and time-consuming to collect.
	As a promising alternative, domain adaptation methods, which adapt the labeled data and model available from an existing source domain to a new target domain, has attracted ever-increasing attention.

	\begin{figure}[tb]
		\centering
		\includegraphics[width=\linewidth]{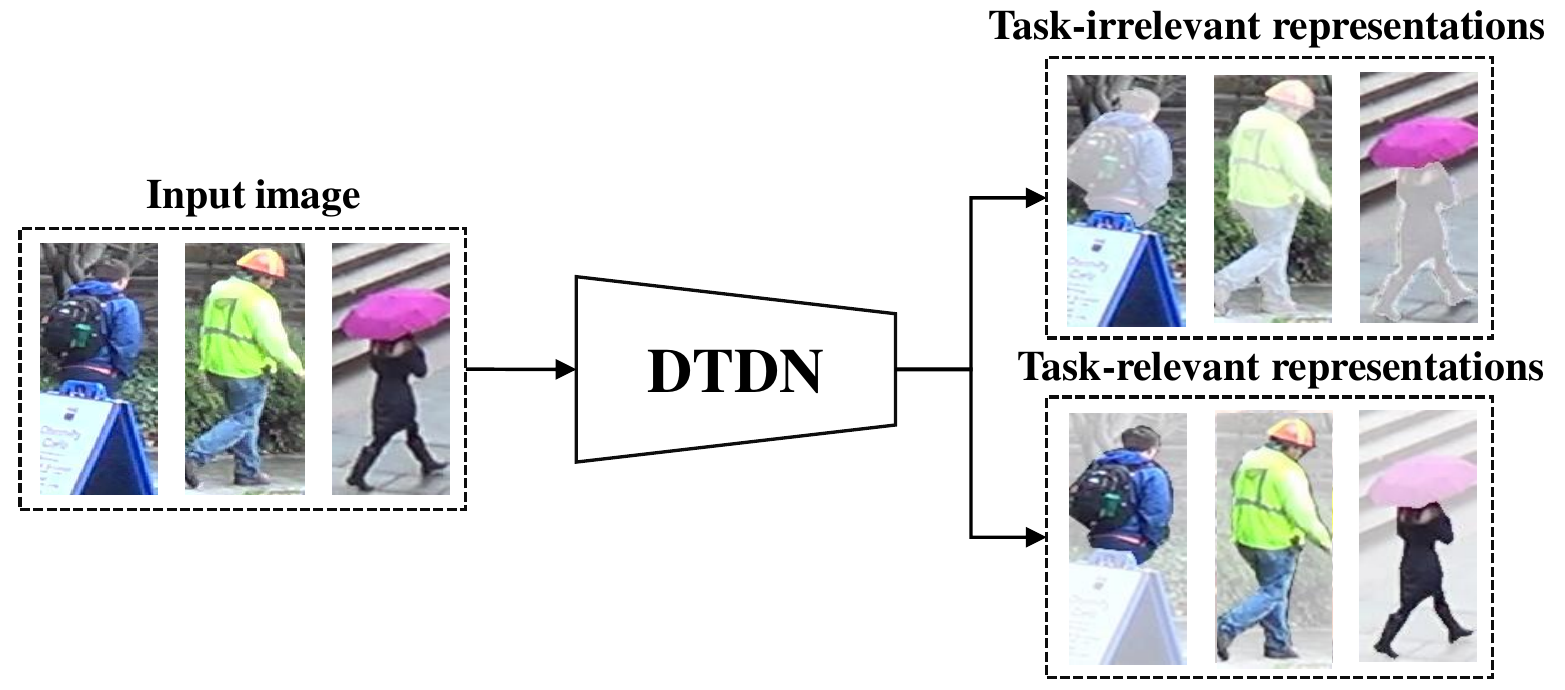}
		\caption{Illustration of DTDN. DTDN disentangles representations into two components (highlight with colors). In a Person-reID task that aims to retrieve images with the same identity, task-irrelevant representations include (1) non-class-discriminative elements, such as information without specific semantics (\emph{e.g.}, lights, colors) and meaningful objects (e.g., road sign, trees, stairs), and (2) non-transferable details (\emph{e.g.}, umbrella). (Best viewed in color)}
		\label{fig1}
	\end{figure}

	Existing endeavors well advocate domain adaptation methods which either search for a feature mapping between the source and target domains~\cite{conf/cvpr/RussoCTC18}, or bridge the domain shift by learning a shared latent space~\cite{conf/icml/LongC0J15,journals/corr/TzengHZSD14}.
	Adversarial domain adaptation methods have also shown promising results~\cite{conf/nips/SnellSZ17,conf/nips/MotiianJID17}.
	Another solution relies on disentanglement, which learns interpretable representations describing domain-invariant information.~\cite{journals/pami/BengioCV13}.

	Despite considerable progress, \blue{the above methods still fail to learn task-oriented information that is both class-discriminative and domain-invariant. \red{Admittedly}, the learned domain-invariant features are not always class-discriminative, such as the same background elements in different domains. Thereby such methods are indeed limited in distinguishing different classes.}
	Moreover, existing methods mainly focus on close-set\footnote{The source and target domains share an identical label space.} settings, where the source and target domains share an identical (or partially identical) label space. 
	They did not fully address the domain adaptation problem in open-set\footnote{The source and target domains have different label spaces.} settings, for instance the image retrieval tasks, where all labels in the target domain are different from those in the source domain and even the number of labels are unknown.

	In this paper, we propose a dynamic task-oriented disentangling network (DTDN) to perform feature disentanglement and domain adaptation in an end-to-end fashion. We deal with both close-set and open-set tasks, notably the complicated open-set image retrieval tasks.
	By breaking the concept of task-orientation into task-relevance and task-irrelevance, DTDN disentangles the feature representations learned by one encoder into task-relevant and task-irrelevant components.
	\blue{As another innovation, we learn only one encoder, which differs from the existing two-encoder learning settings \cite{conf/eccv/LeeTHSY18, conf/nips/BousmalisTSKE16} and is quantitatively shown to implicitly preserves meaningful information.}

	Specifically, task-relevant features contribute to both source and target domains and benefit the task itself, while task-irrelevant features indicate disturbance within a respective domain that should be removed from domain adaptation.
	Fig.~\ref{fig1} illustrates the idea of DTDN. We enforce the disentangled features to describe task-oriented information as specified above. \blue{Since the task-oriented information tackles domain adaptation directly, it can naturally benefit the subsequent tasks.} We conduct extensive experiments to show that the proposed model can generalize well and handle various kinds of tasks, including open-set retrieval (\emph{e.g.}, Person-reID and Vehicle-reID), open-set classification (\emph{e.g.}, Office-31) and close-set classification (\emph{e.g.}, Digit).

	The contributions of this paper are as follows:
	\begin{itemize}
		\item We propose to learn task-oriented representations for unsupervised domain adaptation. By breaking the concept of task-orientation into task-relevance and task-irrelevance, we disentangle the learned representations into task-relevant and task-irrelevant ones.
		\item We introduce DTDN that contains a dynamic network to fulfill representation disentangling in different tasks. DTDN is highly efficient as it largely eliminates the cumbersome settings of complex hyper-parameters.
		\item The proposed model shows significant gains in both close-set and open-set adaptions in the unsupervised setting, especially for the complicated open-set image retrieval tasks. Our unsupervised approach also outperforms a serial of cutting-edge \emph{supervised} ones in Vehicle-reID tasks.
	\end{itemize}

	\section{Related Work}
	
	\textbf{Unsupervised Domain Adaptation.}
	Domain adaptation aims to address the domain-shift problem by generalizing a learner across domains with different distributions~\cite{conf/eccv/SaenkoKFD10,journals/tnn/PanTKY11,conf/icassp/XuLX13,conf/icml/GongGS13,conf/icml/ZhangSMW13}.
	Ben-David \emph{et al.}\ provided upper bounds for a domain-adapted classifier in the target domain, which was later extended to handle multi-source domain adaption~\cite{journals/ml/Ben-DavidBCKPV10}.
	%
	%
	In~\cite{conf/icml/GaninL15,conf/iccv/TzengHDS15,conf/icml/LongC0J15}, deep models are adapted among domains via minimizing the deviation between the source and target distributions to learn transferable representations.
	%
	%
	%
	Specifically, a DANN architecture is introduced to minimize the Maximum Mean Discrepancy (MMD) metric for each domain~\cite{journals/corr/TzengHZSD14}.
	%
	
	Adversarial learning has been recently explored in domain adaptation.
	PixelDA~\cite{conf/cvpr/BousmalisSDEK17} uses adversarial networks to directly produce target images from the source domain, and then perform adaption in the transferred space.
	Hoffman \textit{et al.}~\cite{conf/icml/HoffmanTPZISED18} proposed CyCADA to adapt representations at both pixel and feature levels by pixel cycle consistency with semantic losses.
	However, these methods assume that labels in the target domain are included in the source domain, referred as the close-set setting.
	Thus in the open-set scenario where labels between domains are different, these methods fail to learn discriminative and task-oriented features.

	\textbf{Disentangled Representation.}
	Learning disentangled representation is to decouple the factors of variation~\cite{journals/pami/BengioCV13}.
	One of the first attempts is the bi-linear model which separates the content and style in the underlying set of observations~\cite{conf/nips/TenenbaumF96}.
	Recent efforts resort to generative adversarial networks (GANs)~\cite{conf/nips/GoodfellowPMXWOCB14} and variational autoencoders (VAEs)~\cite{journals/corr/KingmaW13} to learn disentangled representations.
	In a supervised setting, the auxiliary classifier GAN (AC-GAN) was proposed to conduct representation disentanglement~\cite{conf/icml/OdenaOS17}.
	%
	%
	The information-maximizing GAN (InfoGAN) optimizes a lower bound by maximizing the mutual information between latent variables and data variation for disentanglement~\cite{conf/nips/ChenCDHSSA16}.
	However, the above approaches focus on learning disentangled representations in a single domain, which cannot be effectively exploited between different domains.
	Towards cross-domain disentanglement, both domain-specific and domain-shared factors should be identified~\cite{conf/nips/Gonzalez-Garcia18}.
	Some unsupervised methods~\cite{conf/eccv/LeeTHSY18} decouple images into domain-invariant and domain-specific representations to produce diverse image-to-image translation results.
	In contrast to existing works, we take a different focus on the \emph{task} rather than \emph{domain}.
	That is, we aim to separate the learned representations as task-relevant or task-irrelevant, rather than domain-relevant or domain-irrelevant.
	\begin{figure*}[t]
		\begin{center}
			\includegraphics[width=\linewidth]{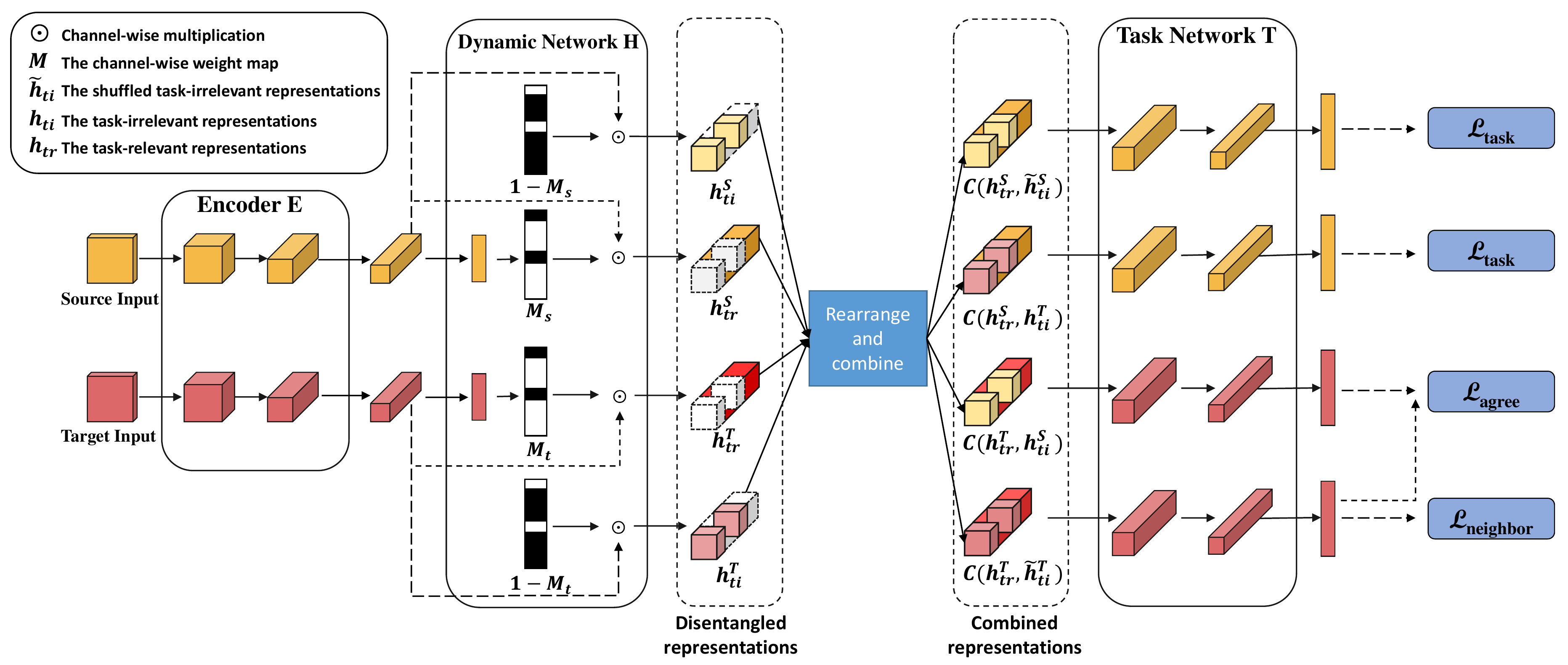}
		\end{center}
		\caption{A conventional encoder (E) projects images into a joint representation space. A dynamic mapping network (H) then divides the resulting representations into two parts, \emph{i.e.}, task-relevant and task-irrelevant ones. These disentangled representations are then shuffled and swapped to capture task-oriented information and bridge the gaps between domains. A task-specific network (T) then guides the overall adaptation task by utilizing the learned representations.}
		\label{fig2}
		\vspace{-1em}
	\end{figure*}

	In terms of the network design in UDA, most works use multiple encoders to disentangle features. For instance, \cite{conf/eccv/LeeTHSY18} embeds all images from both domains into a domain-invariant content space and a domain-specific attribute space using two content and two attribute encoders, respectively. Authors in \cite{conf/nips/BousmalisTSKE16} adopted a shared encoder to capture representations shared among domains, and adopted two private encoders (one for each domain) to capture domain-specific representation. Our one-encoder setting differs from \cite{Liu2018A} that using only one unified encoder in the sense that the representation generated from our encoder is then disentangled into two parts with a dynamically generated mask. Notably, all the above methods resort to decoders or generative models to reconstruct either images or representations. In contrast, our method avoids these steps and reduces the computational cost when using the same backbone network.

	\section{The Proposed Method}
	\label{approach}
	
	We focus on unsupervised domain adaptation. We assume that the source and target domains follow different distributions. Therefore the model trained from the source domain hardly perform well in the target domain.
	Since the two domains with different distributions are still applied to the same task, we design a dynamic task-oriented disentangling network (DTDN) to disentangle the features into task-relevant and task-irrelevant ones in all settings.

	We denote the labeled dataset with $N_s$ samples from the source domain as $\mathbf{X}_s=\{(\mathbf{x}_i^s,y_i^s)\}_{i=0}^{N_s}$, where $\mathbf{x}_i^s\sim{\mathcal{X}_s}$ and label $y_i^s\sim{\mathcal{Y}_s}$.
	Similarly, we denote the unlabeled dataset with $N_t$ samples from the target domain as $\mathbf{X}_t=\{(\mathbf{x}_i^t)\}_{i=0}^{N_t}$, where $\mathbf{x}_i^t\sim{\mathcal{X}_t}$.
	The function $E(\mathbf{x};\bm{\theta}_e)$ parameterized by $\bm{\theta}_e$ maps an image $\mathbf{x}$ to a hidden representation $\mathbf{h}$.
	$H(\mathbf{h};\bm{\theta}_h)$ is a function parameterized by $\bm{\theta}_h$, which disentangles $\mathbf{h}$ into task-relevant $\mathbf{h}_{tr}$ and task-irrelevant $\mathbf{h}_{ti}$.
	A task-specific function $T(\mathbf{r};\bm{\theta}_t)$ parameterized by $\bm{\theta}_t$ converts the rearranged representation $\mathbf{r}$ and combined them into the task-specific predictions $\hat{y}$.
	As shown in Fig.~\ref{fig2}, the proposed DTDN consists of three components: a conventional encoder $E$ with the function $E(\mathbf{x};\bm{\theta}_e)$, a dynamic network $H$ with the function $H(\mathbf{h};\bm{\theta}_{h})$, and a task network $T$ with the function $T(\mathbf{r};\bm{\theta}_t)$.
	Our goal is to explicitly disentangle features extracted from different domains into task-relevant and task-irrelevant components. Consequently, the class-discriminative and transferable task-relevant features are learned to ensure correct predictions in the target domain during the test phase.
	%
	\subsection{Disentangle Representation}
	\label{sec:dis}

	\textbf{Encoder.}
	DTDN conducts disentangling using a shared encoder $E$ trained in both domains.
	$E$ follows a common CNN architecture and shares weights to capture joint representations across domains.
	It maps an image $\mathbf{x}$ to a hidden representation $\mathbf{h}$ with the function $E(\mathbf{x};\bm{\theta}_e)$, where the last few layers are extended to the task network.
	%
	
	%
	\textbf{Dynamic Disentangling Network.}
	To explicitly disentangle features, the ratio of these two features may vary among different tasks.
	To handle this, most disentangling methods split features manually or empirically, which cannot guarantee to obtain the best disentangled results. 
	
	To this end, we propose a dynamic network by adopting an attention model to search a channel-wise weight map $\mathbf{M}=H(\mathbf{h})$, which is then used to disentangle the hidden representation $\mathbf{h}$ as: 
	\begin{equation}
	\begin{aligned}
	\mathbf{h}_{tr}&=\mathbf{M}\odot{E(\mathbf{x};\bm{\theta}_e)}, \\
	\mathbf{h}_{ti}&=(1-\mathbf{M})\odot{E(\mathbf{x};\bm{\theta}_e)},
	\end{aligned}
	\label{equ:attention}
	\end{equation}
	%
	where $\mathbf{h}_{tr}$ and $\mathbf{h}_{ti}$ denote task-relevant and task-irrelevant representations, respectively.
	$\mathbf{M}$ is mapped to continuous values between 0 and 1 by a Sigmoid function, which makes the network differentiable.
	Note that a fully-activated weight map (all ones) means the entire feature would be regarded as task-relevant, the model degrades to the direct transfer from the source domain to the target domain.
	Once the weight map is fully-suppressed (all zeros), the entire feature would be regarded as task-irrelevant, which would crush the following network.

	\textbf{Rearrangement.}
	To capture task-oriented information, given a task-relevant $\mathbf{h}_{tr}$ and a task-irrelevant $\mathbf{h}_{ti}$ from two domains, we rearrange these features with shuffling and swapping operations.
	Firstly, task-irrelevant features are shuffled by switching among different examples from the same domain within a mini-batch, which explicitly breaks their connections to the corresponding ground truth.
	The shuffled features $\tilde{\mathbf{h}}_{ti}^S$ and $\tilde{\mathbf{h}}_{ti}^T$ are then combined with task-relevant ones from the source and target domains respectively as $C(\mathbf{h}_{tr}^S, \tilde{\mathbf{h}}_{ti}^S)$ and $C(\mathbf{h}_{tr}^T, \tilde{\mathbf{h}}_{ti}^T)$.
	Secondly, unshuffled task-irrelevant features are swapped between domains and combined with task-relevant ones as $C(\mathbf{h}_{tr}^S, \mathbf{h}_{ti}^T)$ and $C(\mathbf{h}_{tr}^T, \mathbf{h}_{ti}^S)$.
	The combination is defined as an element-wise addition.
	With the above rearrangement operations, the resulting features should capture information from both domains and encourage disentanglement of task relevance.
	\subsection{Task Network}

	\textbf{Task-oriented Constraint.}
	To exploit label information in the source domain, the rearranged features $C(\mathbf{h}_{{tr}_{i}}^S, \tilde{\mathbf{h}}_{{ti}_{j}}^S)$ and $C(\mathbf{h}_{{tr}_{i}}^S, \mathbf{h}_{{ti}_{k}}^T)$ are assigned the same label as $\mathbf{h}_{{tr}_{i}}^S$, where $i,j,k$ denote different examples within a mini-batch.
	Note that the batch-wise shuffling of $\mathbf{h}_{ti}^S$ will cause misalignment with their assigned labels.
	Also, features $\mathbf{h}_{ti}^T$ from the target domain are not shuffled, as their labels are unknown.
	
	%
	To make the model predict labels that for a given task, a task-specific constraint is defined as
	%
	\begin{equation}
	\begin{split}
	\label{equ:task-oriented}
	\mathcal{L}_{task}=
	&\mathbb{E}_{\mathbf{h}^S\sim{\mathbf{H}}^S,\mathbf{h}^T\sim{\mathbf{H}}^T}\bigg[J\Big (T \big (C( \mathbf{h}_{tr_{i}}^S, \tilde{\mathbf{h}}_{ti_{j}}^S) \big ), y_i^S \Big) \\
	&+J\Big (T \big (C( \mathbf{h}_{tr_{i}}^S, \mathbf{h}_{ti_{k}}^T) \big ), y_i^S \Big )\bigg] + w\big|\big|\mathbf{M}^S\big|\big|_1,
	\end{split}
	\end{equation}
	where $J(\cdot, \cdot)$ is the cross-entropy loss function and $y_i^S$ denotes the label in the source domain.
	%
	%
	$\mathcal{L}_{task}$ is then minimized to eliminate task-relevant information within $\tilde{\mathbf{h}}_{ti}^S$ and $\mathbf{h}_{ti}^T$, which should facilitate the disentangling for the task. We use an $\ell_1$ regularization to avoid a fully-activated weight map. And $w$ is set to $10^{-4}$ in all experiments.
	
	\textbf{Unsupervised Agreement Constraint.}
	Though there is no label in the target domain, we use the task network to predict labels in the target domain close to the ground truth.
	A pseudo label $\tilde{y}$ is assigned to each image $\mathbf{x}$ of the target domain based on its index in the dataset, such that each image is considered as a unique class.
	To only take samples from the target domain into account, $\mathbf{h}_{ti_{j}}^S$ that comes from the source domain is not shuffled.
	$C(\mathbf{h}_{tr_{i}}^T, \mathbf{h}_{ti_{j}}^S)$ and $C(\mathbf{h}_{tr_{i}}^T, \tilde{\mathbf{h}}_{ti_{k}}^T)$ are therefore assigned the same label as $\mathbf{h}_{tr_{i}}^T$.
	%
	$\mathcal{L}_{agree}$ is then defined based on the $\ell_1$ distance among predictions of hidden representations as
	\begin{equation}
	\small
	\begin{split}
	\label{equ:agreement}
	\mathcal{L}_{agree}=
	&\mathbb{E}_{\mathbf{h}^S\sim{\mathbf{H}}^S,\mathbf{h}^T\sim{\mathbf{H}}^T}\bigg[
	\sum_{i=0}^{N_{T}}\Big|
	P\Big(\tilde{y}_i\big|T\big(C(\mathbf{h}_{tr_{i}}^T, \mathbf{h}_{ti_{j}}^S)\big)\Big)\\
	&-
	P\Big(\tilde{y}_i\big|T\big(C(\mathbf{h}_{tr_{i}}^T, \tilde{\mathbf{h}}_{ti_{k}}^T)\big)\Big)\Big|\bigg] + w\big|\big|\mathbf{M}^T\big|\big|_1,
	\end{split}
	\end{equation}
	where $N_{T}$ denotes the number of pseudo labels.
	Minimizing $\mathcal{L}_{agree}$ further increases the task-irrelevance of $\mathbf{h}_{ti}^S$ and $\tilde{\mathbf{h}}_{ti}^T$ to facilitate the disentanglement. Similar to Eqn.~\ref{equ:task-oriented}, an $\ell_1$ regularization is used to avoid a mask with all ones.
	%
	
	\textbf{Target Domain Alignment.}
	Since the above two constraints only guarantee the irrelevance of task-irrelevant representations rather than the relevance of the task-relevant ones, \emph{relevant information} that only contributes to the source domain may still exist. Therefore, an anchor neighborhood discovery method \cite{journals/corr/abs-1904-11567} is adopted to align the target domain, which helps remove such information from task-relevant components, as its existence would break the alignment.
	%
	%
	To explain, a matrix $\mathbf{A}$ is initialized with shape $N\times{D}$, where $N$ and $D$ denote the size of the target training set and the feature dimensions, respectively.
	For each training iteration, $\mathbf{A}$ is updated with newly computed feature maps from the task network with an update rate $\eta$: 
	%
	\begin{equation}
	\label{equ:updateTarget}
	\mathbf{A}_i^{(n+1)} = (1 - \eta)\cdot \mathbf{A}_i^{(n)} + \eta\cdot T \big (C( \mathbf{h}_{tr_{i}}^T, \tilde{\mathbf{h}}_{ti_{k}}^T) \big ),
	\end{equation}
	where $\mathbf{A}_i^{(n)}$ and $\mathbf{A}_i^{(n+1)}$ denote the $i$-th row of the matrix $\mathbf{A}$ before and after the update, respectively.
	
	When the task is close-set or simple open-set with a known number of labels in the target domain, we first predict each image's label, and then update the matrix based on the location of the top-$k$ most similar images.
	Therefore, the anchor neighborhood discovery loss is defined as
	\begin{equation}
	\label{equ:close}
	\mathcal{L}_{neighbor} = -\sum^{N_{bs}}_{i=1}\log(\sum_{j\in \mathcal{K}_i}d_{i,j}),
	\end{equation}
	where $N_{bs}$ denotes the training mini-batch size, and $d_{i,j}$ denotes the Cosine similarity, defined as
	\begin{equation}
	\label{equ:cos}
	d_{i,j} = \frac{\exp(\mathbf{A}^T_i\mathbf{A}_j/\tau)}{\sum^N_{n=1}\exp(\mathbf{A}^T_i\mathbf{A}_n/\tau)},
	\end{equation}
	where a temperature parameter $\tau$ is used to control the distribution concentration degree~\cite{journals/corr/HintonVD15}.

	For the complicated open-set task where a specific amount of labels remains unknown, we locate the top-$k$ most similar images of $\mathbf{x}$ as $\mathcal{K}_\mathbf{x}$ in matrix $\mathbf{A}$ by computing the pair-wise Cosine similarities of each image $\mathbf{x}$ as Eqn.~\ref{equ:cos}, and minimize their distances as Eqn.~\ref{equ:close}.
	%
	
	\subsection{Objective Function}
	Given an input image $\mathbf{x}$ and a prediction $\hat{y}$, the training of DTDN is to minimize the loss $\mathcal{L}(E,H,T)$ \emph{w.r.t.} parameters $\Theta=\{\bm{\theta}_e,\bm{\theta}_{h},\bm{\theta}_t\}$ as
	\begin{equation}
	\label{equ:Objective}
	\mathcal{L}(E,H,T)=\mathcal{L}_{task}+\lambda_1\cdot\mathcal{L}_{neighbor}+\lambda_2\cdot\mathcal{L}_{agree},
	\end{equation}
	where $\lambda_1$ and $\lambda_2$ control interaction of different loss terms.

	We set up the above three loss functions to learn task-relevant features that are shared and transferable across domains. To be more specific, $L_{task}$ and $L_{agree}$ guide the network to preserve information critical to the task with task-relevant features, \emph{i.e.}, by minimizing classification errors and disturbing task-irrelevant ones. Similarly, $L_{task}$ and $L_{neighbor}$ force the network to learn transferable task-relevant information. To explain, information only contributing to the task in the source (\emph{resp.} target) domain will result in an increased $L_{neighbor}$ (\emph{resp.} $L_{task}$), if it’s incorrectly divided into the task-irrelevant parts.
	
	\section{Experiments}
	\label{Experiments}
	\blue{We evaluate the proposed model by conducting both the retrieval and classification tasks.}
	\blue{For the retrieval tasks, Person-reID and Vehicle-reID are conducted, which are complicated, open-set problems with no labels shared between the source and target domains. Such problems are challenging in the sense that they involve both task-relevant and task-irrelevant representations.}
	T-SNE visualization also shows the correctness and meaningfulness of disentangled representations. 
	As for the classification tasks, Office-31 classification and digits recognition tasks are chosen for open-set and close-set settings, respectively.
	\subsection{Experimental Details}
	
	\textbf{Datasets of Retrieval Tasks.} We evaluated the proposed model in both the tasks of Person-reID and Vehicle-reID. For Person-reID, two widely used large-scale benchmarks are used: Market-1501 (Market)~\cite{zheng2015scalable} and DukeMTMC-reID (Duke)~\cite{zheng2017unlabeled}. Market has 32,668 person images of 1,501 identities captured from 6 camera views. Duke has 36,411 person images of 1,404 identities captured from 8 camera views.  
	For Vehicle-reID tasks, VeRi~\cite{DBLP:conf/eccv/LiuLMM16} and VehicleID~\cite{liu2016deep} datsets are used. VeRi contains 776 different vehicles with 49,357 images captured from 20 cameras. VehicleID has 21,763 images of 26,297 vehicles. The test set is split into three subsets, \emph{i.e.}, small, medium and large, with 800, 1600 and 2400 different vehicles, respectively. 

	\textbf{Datasets of Classification Tasks.} We also evaluate the proposed model with classification adaptation tasks. 
	For the open-set settings, we choose Office-31~\cite{conf/eccv/SaenkoKFD10} that has 4,652 images of 31 categories from three distinct domains: Amazon (A), Webcam (W), and DSLR (D). We mainly follow the settings of~\cite{DBLP:conf/eccv/SaitoYUH18}, where the target domain has all classes in the source domain and further contains target-specific classes. 
	%
	For the close-set experiments, three digits datasets are used: MNIST (M), USPS (U), and SVHN (S). They have 70,000, 9,298 and 99,289 images, respectively. Three adaptation tasks are conducted: M$\to$U, U$\to$M, and S$\to$M. 
	\begin{table*}[t]
		\caption{Accuracy (\%) on Person-reID for Unsupervised Domain Adaptation}
		\label{table:person-reid}
		\centering
		\resizebox{\textwidth}{!}{
			\begin{tabular}{c|ccccc|ccccc}
				\hline
				\multirow{2}{*}{Methods} & \multicolumn{5}{c|}{DukeMTMC-ReID $\to$ Market-1501} & \multicolumn{5}{c}{Market-1501 $\to$ DukeMTMC-ReID} \\ \cline{2-11}
				& Rank-1   & Rank-5  & Rank-10  & Rank-20  & \emph{m}AP  & Rank-1   & Rank-5  & Rank-10  & Rank-20  & \emph{m}AP  \\ \hline\hline
				LOMO~\cite{liao2015person}                     & 27.2     & 41.6    & 49.1     & -        & 8.0    & 12.3     & 21.3    & 26.6     & -        & 4.8    \\
				Bow~\cite{zheng2015scalable}                      & 35.8     & 52.4    & 60.3     & -        & 14.8   & 17.1     & 28.8    & 34.9     & -        & 8.3    \\ \hline
				ResNet-50~\cite{DBLP:conf/cvpr/HeZRS16}          & 43.1     & 60.8    & 68.1     & 74.7     & 17.0   & 33.1     & 49.3    & 55.6     & 61.9     & 16.7   \\
				SPGAN+LMP~\cite{deng2018image}                & 57.7     & 75.8    & 82.4     & 87.6     & 26.7   & 46.4     & 62.3    & 68.0     & 87.6     & 26.2   \\
				TJ-AIDL~\cite{wang2018transferable}                  & 58.2     & 74.8    & 81.1     & 86.5     & 26.5   & 44.3     & 59.6    & 65.0     & 70.0     & 23.0   \\ 
				HHL~\cite{DBLP:conf/eccv/ZhongZLY18}              &62.2    &78.8    &84.0    &-    &31.4    &46.9    &61.0   &66.7    &-    &27.2 \\
				PAUL\cite{yang2019patch} &66.7&-&-&-&36.8&56.1&-&-&-&35.7\\
				\red{UCDA+SOT} \cite{Qi_2019_ICCV} &73.7&-&-&-&\textbf{49.6}&64.0&-&-&-&45.6\\
				\hline
				DTDN (Ours)                     & 75.2     & 85.4    & 88.9     & 92.0     & 43.9      & 65.4        & 75.6       & 78.5        & 81.6        & 44.9 \\
				\red{DTDN+IN (Ours)}                     & \textbf{80.1}     & \textbf{87.8}    & \textbf{90.8}     & \textbf{93.6}     & 47.1      & \textbf{70.0}        & \textbf{78.9}       & \textbf{82.0}        & \textbf{84.2}        & \textbf{47.7}
				\\ \hline
			\end{tabular}
		}
		\vspace{-1.5em}
	\end{table*}
	\textbf{Details}
	\blue{For each training task, we use all labeled source examples and unlabeled target examples in the training set.}
	Backbone CNNs for the encoder $E$ and task network $T$ of each task are chosen as follows.
	For Person-reID, Vehicle-reID, and Office-31 classification, the backbone network is \emph{ResNet-50}~\cite{DBLP:conf/cvpr/HeZRS16} with an additional 1024-d fully connected layer and the cross-entropy loss. Notably, the last layer's stride is set to 1 to obtain richer information for both reID tasks.
	\emph{ResNet-20} is used for tasks on Digits.
	\red{We also employ an Instance Normalization (IN) \cite{journals/corr/UlyanovVL16} for retirval tasks in the residual bottlenecks at ResNet's layer-3 and refer to it as DTDN+IN.
		Since the dataset for the classification task is small, we did not employ the IN module to avoid overfitting in these tasks.}
	For the dynamic network, two convolution layers followed by a Sigmoid activation function are used to obtain a channel-wise weight map for each image.
	
	For the image retrieval, Office-31 classification, and digits recognition tasks, the model is trained for 40, 19, and 100 epochs, respectively. The batch size is set to 128 for the above tasks. The SGD with Nesterov momentum at 0.9 is used for training, and $\eta$ is set to 0.01 in Eqn.~\ref{equ:updateTarget} for feature update. The temporal temperature parameter $\tau$ in Eqn.~\ref{equ:cos} is set to 0.05 for reID and digits tasks, and 0.5 for Office-31, respectively. $\lambda_1$ is set to 0.8 and $\lambda_2$ is 1 in Eqn.~\ref{equ:Objective}. For target domain alignment, top-6, top-9, and top-500 most similar images are examined, based on the distance calculated using Eqn.~\ref{equ:close} for reID, Office-31 and digits recognition tasks.
	
	\subsection{Qualitative Results of Learned Representations}
	\blue{We evaluate the learned representations by disentangling them into task-relevant and task-irrelevant parts through an encoder $E$ and a dynamic network $H$.} We then use these two representations to retrieve top-$k$ similar images from the test set through a task network $T$, respectively.

\begin{figure}[t]
		\centering
		\subfigure[Task-irrelevance]{
			\label{fig:searchresult-ir}
			\includegraphics[width=0.47\linewidth]{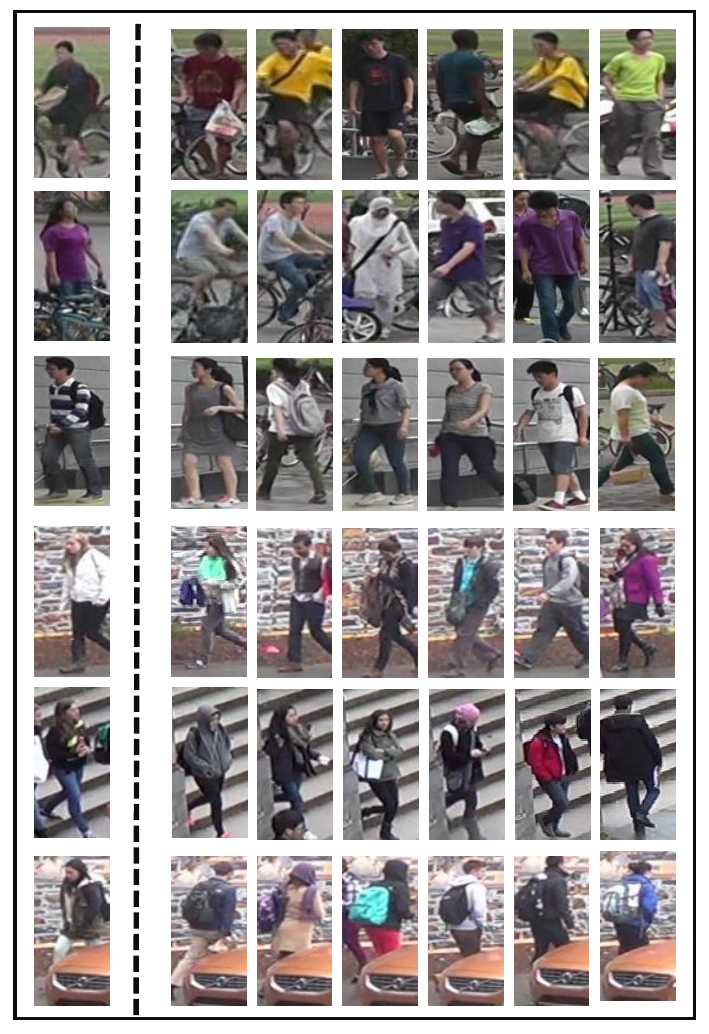}
		}
		\subfigure[Task-relevance]{
			\label{fig:searchresult-r}
			\includegraphics[width=0.47\linewidth]{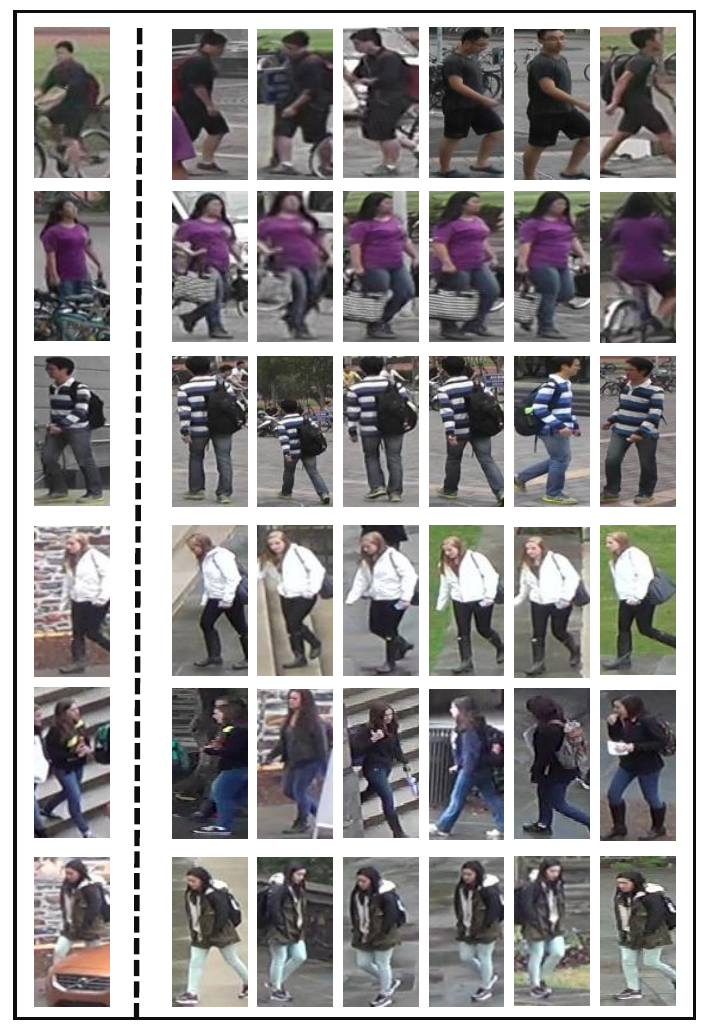}
		}
		\caption{Retrieval results with only task-irrelevant or task-relevant representations for Person-reID}
		\label{fig:searchresult}
		\vspace{-1.5em}
	\end{figure}

\begin{figure}[t]
		\centering
		\subfigure[Task-irrelevance]{
			\label{fig:vehicle-ti}
			\includegraphics[width=0.47\linewidth]{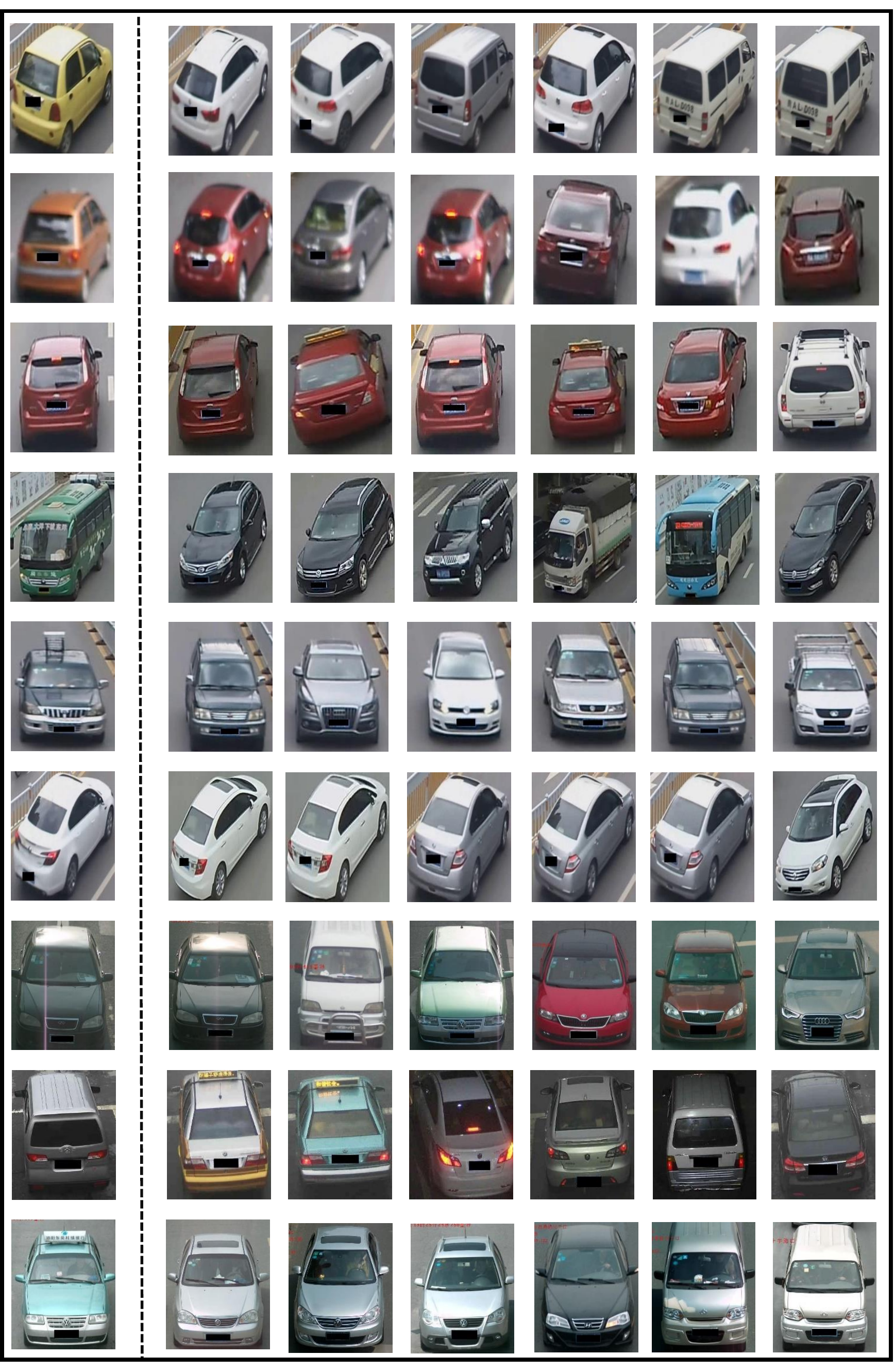}
		}
		\subfigure[Task-relevance]{
			\label{fig:vehicle-tr}
			\includegraphics[width=0.47\linewidth]{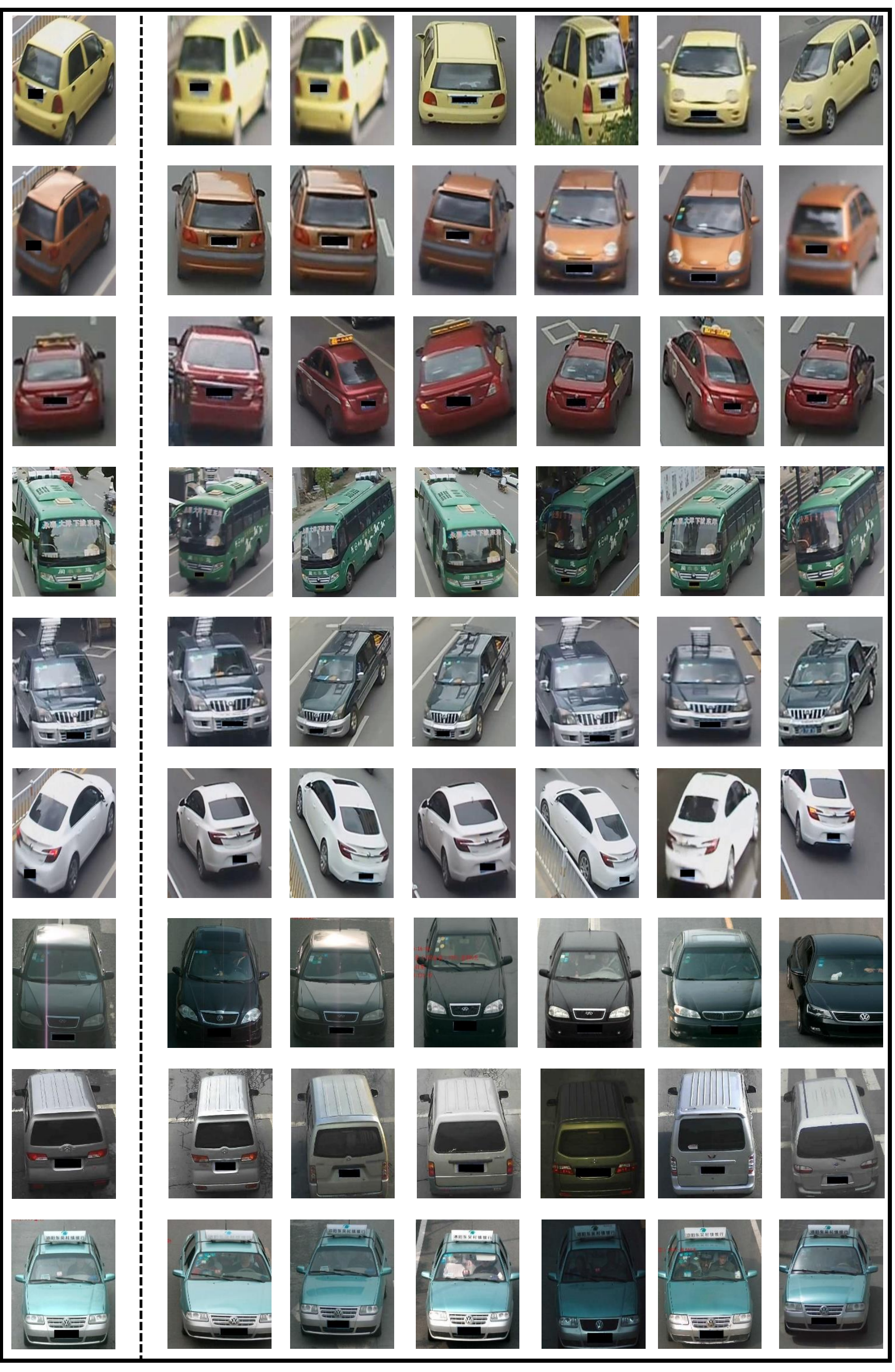}
		}
		\caption{Retrieval results with only task-irrelevant or task-relevant representations for Vehicle-reID}
		\label{fig:searchresult-ve}
		\vspace{-1.5em}
	\end{figure}

	\textbf{Task-irrelevant Representations. }Fig.\ \ref{fig:searchresult-ir} and Fig.\ \ref{fig:vehicle-ti} present images retrieved with only task-irrelevant representations in Person-reID and Vehicle-reID, respectively. For Person-reID, all top-6 retrieved images contain ``bicycle'' when a bicycle appears in the query-set in rows 1 and 2. Similarly, all images retrieved in this way contain similar details that are irrelevant to the reID task, such as the metal barriers, walls and stairs in rows 3-5, and the car in row 6.
	As for Vehicle-reID, the cars in retrieved images clearly have the same angle of views as the queried ones, while those features that are critical to the identification are not concerned at all, such as IDs, models and colors.
	
	The above results indicate that DTDN can correctly assign information that is indeed irrelevant to the task-irrelevant components, such as cars, bicycles, and backgrounds for Person-reID and angle of views for Vehicle-reID. Moreover, meaningful information, as is described above, also appears consistently within retrieval results, which further indicates the effectiveness of our disentanglement method. It highlights the superiority of our disentanglement method over simple noise-removal regularization.
	
	\begin{table*}[!ht]
		\caption{Accuracy (\%) on Vehicle-reID for Unsupervised Domain Adaptation. (All compared methods are \emph{supervised} learning)}
		\label{table:vehicle-reid}
		\resizebox{\textwidth}{!}{
			\begin{tabular}{c|cccc|c|ccc|ccc|ccc}
				\hline
				\multicolumn{5}{c|}{VehicleID $\to$ VeRi} & \multicolumn{10}{c}{VeRi $\to$ VehicleID} \\ \hline\hline
				Settings & \multicolumn{4}{c|}{Query=1678, Test=11579} & Settings & \multicolumn{3}{c|}{Test Size = 800} & \multicolumn{3}{c|}{Test Size = 1600} & \multicolumn{3}{c}{Test Size = 2400} \\ \hline
				Methods & \emph{m}AP & Rank-1 & Rank-5 & Rank-20 & Methods & Rank-1 & Rank-5 & Rank-20 & Rank-1 & Rank-5 & Rank-20 & Rank-1 & Rank-5 & Rank-20 \\ 
				\hline
				LOMO~\cite{liao2015person}          & 9.8  & 23.9 & 39.1 & 54.5 & LOMO~\cite{liao2015person}          & 19.8 & 32.0 & 45.0 & 18.9 & 29.2 & 39.9 & 15.3 & 25.3 & 36.0 \\
				DGD~\cite{xiao2016learning}           & 17.9 & 50.7 & 67.5 & 79.9 & DGD~\cite{xiao2016learning}           & 44.8 & 66.3 & 81.5 & 40.3 & 65.3 & 76.8 & 37.3 & 57.8 & 70.3 \\
				FACT~\cite{liu2016deep}          & 18.7 & 51.9 & 67.2 & 79.6 & FACT~\cite{liu2016deep}          & 49.5 & 68.1 & 78.5 & 44.6 & 64.6 & 75.3 & 39.9 & 60.3 & 72.9 \\
				ResNet-50~\cite{DBLP:conf/cvpr/HeZRS16}  & 21.6 & 64.1 & 74.4 & 84.0 & ResNet-50~\cite{DBLP:conf/cvpr/HeZRS16}      & 45.4  & 64.2 & 77.4 & 42.0  & 60.1 & 72.9 & 36.3  & 53.3 & 67.5 \\
				XVGAN~\cite{DBLP:conf/bmvc/Zhou017}         & 24.7 & 60.2 & 77.0 & 88.1 & XVGAN~\cite{DBLP:conf/bmvc/Zhou017}         & 52.9 & \textbf{80.9} & \textbf{91.9} & 49.6& \textbf{71.4} & \textbf{81.7} & 44.9 & \textbf{66.7} & \textbf{78.0} \\
				SiameseVisual~\cite{shen2017learning} & 29.5 & 41.1 & 60.3 & 79.9 & VGG+CCL~\cite{liu2016deep}       & 43.6 & 64.8 & 80.1 & 39.9 & 63.0 & 76.1 & 35.7 & 56.2 & 68.4 \\
				OIFE~\cite{wang2017orientation}          & \textbf{48.0} & 65.9 & \textbf{87.7} & \textbf{96.6} & MixedDiff+CCL~\cite{liu2016deep} & 48.9 & 75.7 & 88.5 & 45.1 & 68.9 & 79.9 & 41.1 & 63.4 & 76.6 \\ 
				\hline
				DTDN (Ours)   & 30.1  & 74.5  & 83.2  & 86.1 & DTDN (Ours)  & 52.2  & 65.1  & 74.8    & 49.6    & 65.4     & 73.4   & 44.3  & 60.1     & 71.9 \\
				DTDN+IN (Ours)   & 31.3  & \textbf{75.3}  & 85.0  & 86.2 & DTDN+IN (Ours) & \textbf{52.8}  & 65.3  & 75.3    & \textbf{50.7}    & 65.2     & 72.7   & \textbf{45.7}  & 59.6     & 73.9 \\
				\hline
			\end{tabular}
		}
		\vspace{-1em}
	\end{table*}
	
	\textbf{Task-relevant Representations.}
	Fig.~\ref{fig:searchresult-r} and Fig.~\ref{fig:vehicle-tr} present images retrieved with only task-relevant representations in Person-reID and Vehicle-reID, respectively. \blue{For Person-reID, the persons in retrieved images have clothes with the same color, hair with the similar length, and the same gender as the queried one.}
	Similarly, the cars in retrieved images for Vehicle-reID have the same color and shape as the queried one. Notably, although these images are captured in different environments, the learned features that are critical to the identification remain the same.
	
	%
	%
	\blue{These results also demonstrate that DTDN can \red{adaptively} disentangle the representation into two components}: the task-relevant ones that benefit the task, \emph{e.g.}, people's appearance characters, and the task-irrelevant ones that do not contribute to both domains for a given task, \emph{e.g.}, domain-specific background information.
	
\begin{figure}[ht]
		\vspace{-1em}
		\subfigure[Total representation]{
			\label{fig:filter1}
			\includegraphics[width=0.55\linewidth]{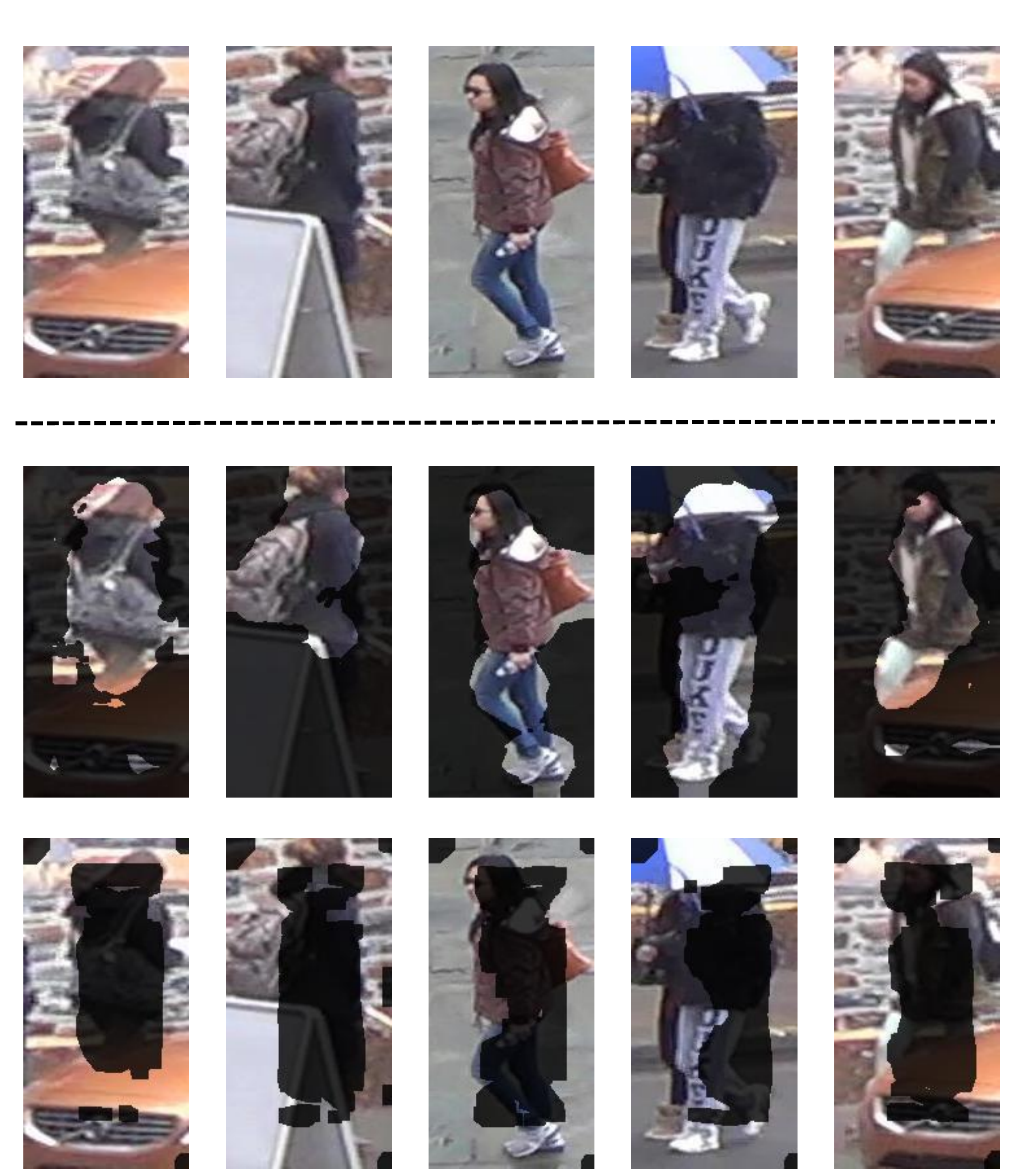}
		}
		\subfigure[Individual filter]{
			\label{fig:filter2}
			\includegraphics[width=0.38\linewidth]{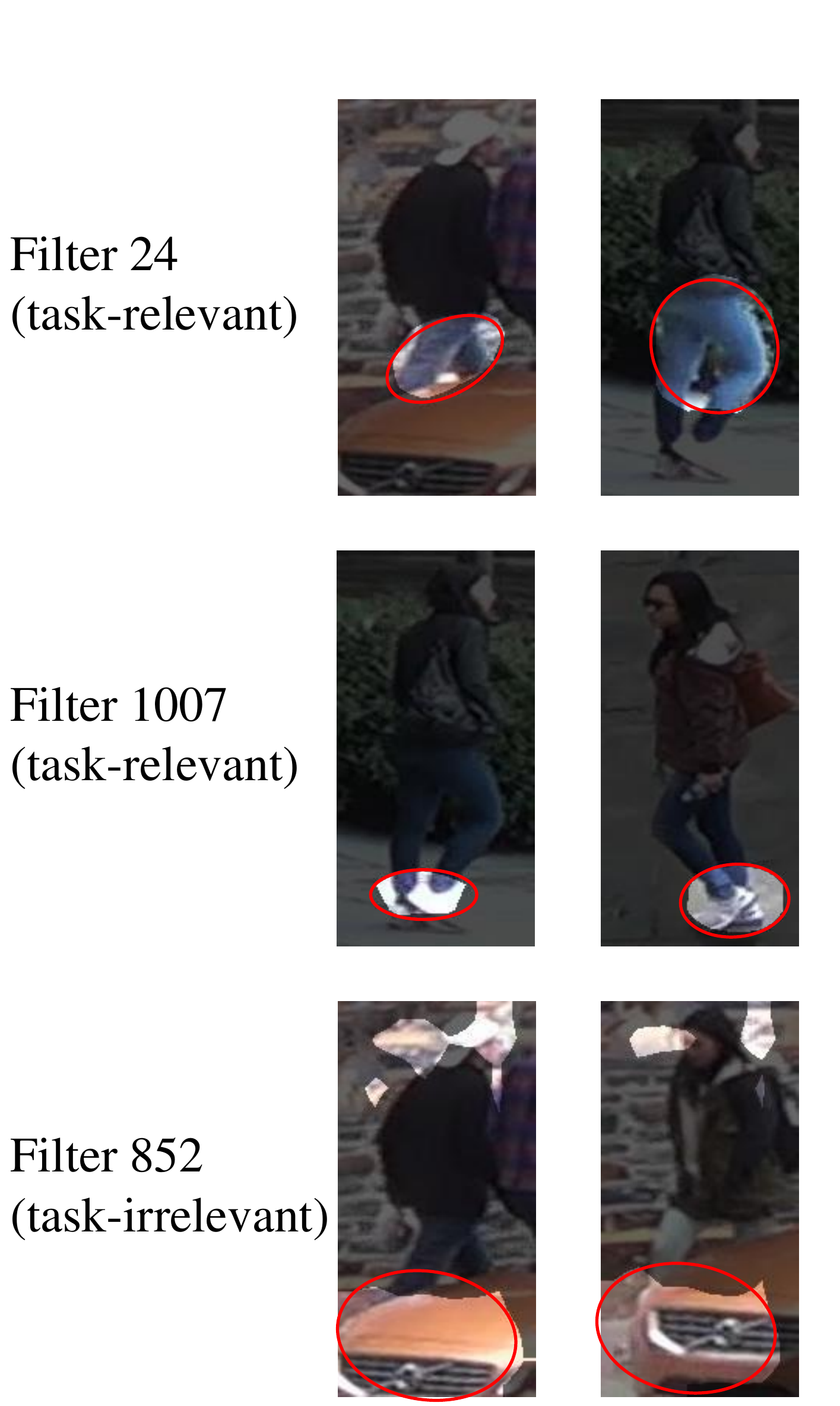}
		}
		\caption{\red{Visualization of the receptive field of representation. (a) Original images, task-relevant and task-irrelevant representations. (b) Results of filter 24, 1007 and 852, respectively.}}
		\vspace{-1em}
	\end{figure}

	\textbf{Receptive field.}
	Following the same method described in \cite{conf/cvpr/BauZKO017}, we compute the response values and areas of each channel in the task-relevant and task-irrelevant representations that we obtained. The responses along all channels (2048 for Person-reID) are then added up to show the receptive field of total representations. Fig.~\ref{fig:filter1} shows receptive field of representations from the DukeMTMC-reID.
	We can observe that the response areas for task-relevant representations mainly focus on the person rather than background as shown in the first two rows, such as cars, block walls and road sign. While for task-irrelevant representations (the last row), \blue{since they usually contain features about light and colors that \red{fill the entire image}, the response areas are moderately large.} Despite that, task-irrelevant representations also respond selectively to elements like cars and walls that are indeed irrelevant to the task, yet refuse to respond to most critical areas that are relevant to the Person-reID task.
	
	Moreover, Fig.~\ref{fig:filter2} shows the response areas of selected individual filters. The filters 24 and 1007 of task-relevant representation respond to legs and shoes, respectively, while the filter 852 of task-irrelevant representations responds to cars at the bottom of the image.

	\begin{table*}[!ht]
		\caption{Rank-1 Accuracy (\%) on Office-31 for UDA using \emph{ResNet-50}}
		\begin{adjustbox}{max width=1.0\linewidth}
			\begin{tabular}{c|c|c|c|c|c|c|c|c|c|c|c|c|c|c}
				\hline
				\multirow{2}{*}{Methods}   &\multicolumn{2}{c|}{A$\to$W} &\multicolumn{2}{c|}{D$\to$W} &\multicolumn{2}{c|}{A$\to$D} &\multicolumn{2}{c|}{W$\to$D} &\multicolumn{2}{c|}{D$\to$A} &\multicolumn{2}{c|}{W$\to$A} 
				&\multicolumn{2}{c}{Avg}\\  \cline{2-15}
				- &OS &OS* &OS &OS* &OS &OS* &OS &OS* &OS &OS* &OS &OS* &OS &OS* \\ \hline\hline
				ResNet-50~\cite{DBLP:conf/cvpr/HeZRS16} &82.5$\pm$1.2 &82.7$\pm$0.9 &94.1$\pm$0.3 &94.3$\pm$0.7 &85.2$\pm$0.3 &85.5$\pm$0.9 &96.6$\pm$0.2 &97.0$\pm$0.4 &71.6$\pm$0.1 &71.5$\pm$1.1 &75.5$\pm$1.0 &75.2$\pm$1.6 &84.2 &84.4 \\ 
				RTN~\cite{DBLP:conf/nips/LongZ0J16} &85.6$\pm$1.2 &88.1$\pm$1.0 &94.8$\pm$0.3 &96.2$\pm$0.7 &89.5$\pm$1.4 &90.1$\pm$1.6 &97.1$\pm$0.2 &98.7$\pm$0.9 &72.3$\pm$0.9 &72.8$\pm$1.5 &73.5$\pm$0.6 &75.9$\pm$1.4 &85.4 &86.8 \\ 
				DANN~\cite{conf/icml/GaninL15}    & 85.3$\pm$0.7 & 87.7$\pm$1.1 & 97.5$\pm$0.2 & 98.3$\pm$0.5 & 86.5$\pm$0.6 & 87.7$\pm$0.6 & 99.5$\pm$0.1 & 100.0$\pm$.0  &75.7$\pm$1.6 & 76.2$\pm$0.9 & 74.9$\pm$1.2 & 75.6$\pm$0.8 &86.6 &87.6               \\
				ATI-$\lambda$~\cite{conf/iccv/BustoG17}   &87.4$\pm$1.5 & 88.9$\pm$1.4 & 93.6$\pm$1.0 & 95.3$\pm$1.0 & 84.3$\pm$1.2 & 86.6$\pm$1.1 & 96.5$\pm$0.9 & 98.7$\pm$0.8 & 78.0$\pm$1.8 & 79.6$\pm$1.5 & 80.4$\pm$1.4 & 81.4$\pm$1.2 &86.7 &88.4       \\
				OSBP~\cite{DBLP:conf/eccv/SaitoYUH18}  &86.5$\pm$2.0 &87.6$\pm$2.1 &88.6$\pm$1.4 &97.0$\pm$1.0 &\textbf{96.5$\pm$0.4} &89.2$\pm$1.3 &97.9$\pm$0.9 &98.7$\pm$0.6  &88.9$\pm$2.5 &90.6$\pm$2.3 &85.8$\pm$2.5 &84.9$\pm$1.3
				&90.8 &91.3 \\
				STA~\cite{DBLP:conf/eccv/SaitoYUH18}  &\textbf{89.5$\pm$0.6} &\textbf{92.1$\pm$0.5} &97.5$\pm$0.2 &96.5$\pm$0.5 &93.7$\pm$1.5 &96.1$\pm$0.4 &\textbf{99.5$\pm$0.2} &99.6$\pm$0.1  &\textbf{89.1$\pm$0.5} &\textbf{93.5$\pm$0.8} &\textbf{87.9$\pm$0.9} &\textbf{87.4$\pm$0.6}  
				&\textbf{92.9} &\textbf{94.1}
				\\ 
				\hline
				DTDN (Ours)    &88.5$\pm$3.6 &91.3$\pm$2.8 &\textbf{99.6$\pm$0.1} &\textbf{99.9$\pm$0.1} &96.3$\pm$1.8 &\textbf{99.2$\pm$0.8} &98.6$\pm$0.2 &\textbf{99.9$\pm$0.1} &85.9$\pm$0.2 &88.0$\pm$0.3 &81.9$\pm$2.2 &85.5$\pm$1.8 &91.8 &94.0   \\  \hline
			\end{tabular}
		\end{adjustbox} 
		\label{table:office31}
		\vspace{-1em}
	\end{table*}
	\begin{table}[ht]
		\caption{Rank-1 Accuracy (\%) on Digits for UDA}
		\begin{adjustbox}{max width=1.0\linewidth}
			\begin{tabular}{c|c|c|c}
				\hline
				Model   &M $\to$ U & U $\to$ M &S $\to$ M \\  \hline\hline
				Supervised Learning &99.22 &99.79 &99.79 \\ 
				Source Only &99.43         &95.17                  &85.16 \\
				\hline
				UNIT~\cite{Liu2017Unsupervised}    &95.97                  &93.85                  &90.53                  \\
				ACGAN~\cite{conf/cvpr/Sankaranarayanan18a}   &92.80                  &90.80                  &92.40                  \\
				UFDN~\cite{Liu2018A}    &97.13                  &93.77                  &95.01                  \\
				GPDA~\cite{journals/corr/abs-1902-08727}    &96.45                  &96.37                  &98.20                  \\
				DIAL~\cite{journals/corr/abs-1811-12751}    &97.60                  &99.12                  &95.85                  \\ \hline
				DTDN (Ours)    &\textbf{99.08}         &\textbf{99.15}         &\textbf{98.15}         \\  \hline
			\end{tabular}
		\end{adjustbox}
		\label{table:digits}
		\vspace{-1em}
	\end{table}

\begin{figure}[ht]
		\subfigure[Domain distributions]{
			\label{fig:tsneL}
			\includegraphics[width=0.47\linewidth]{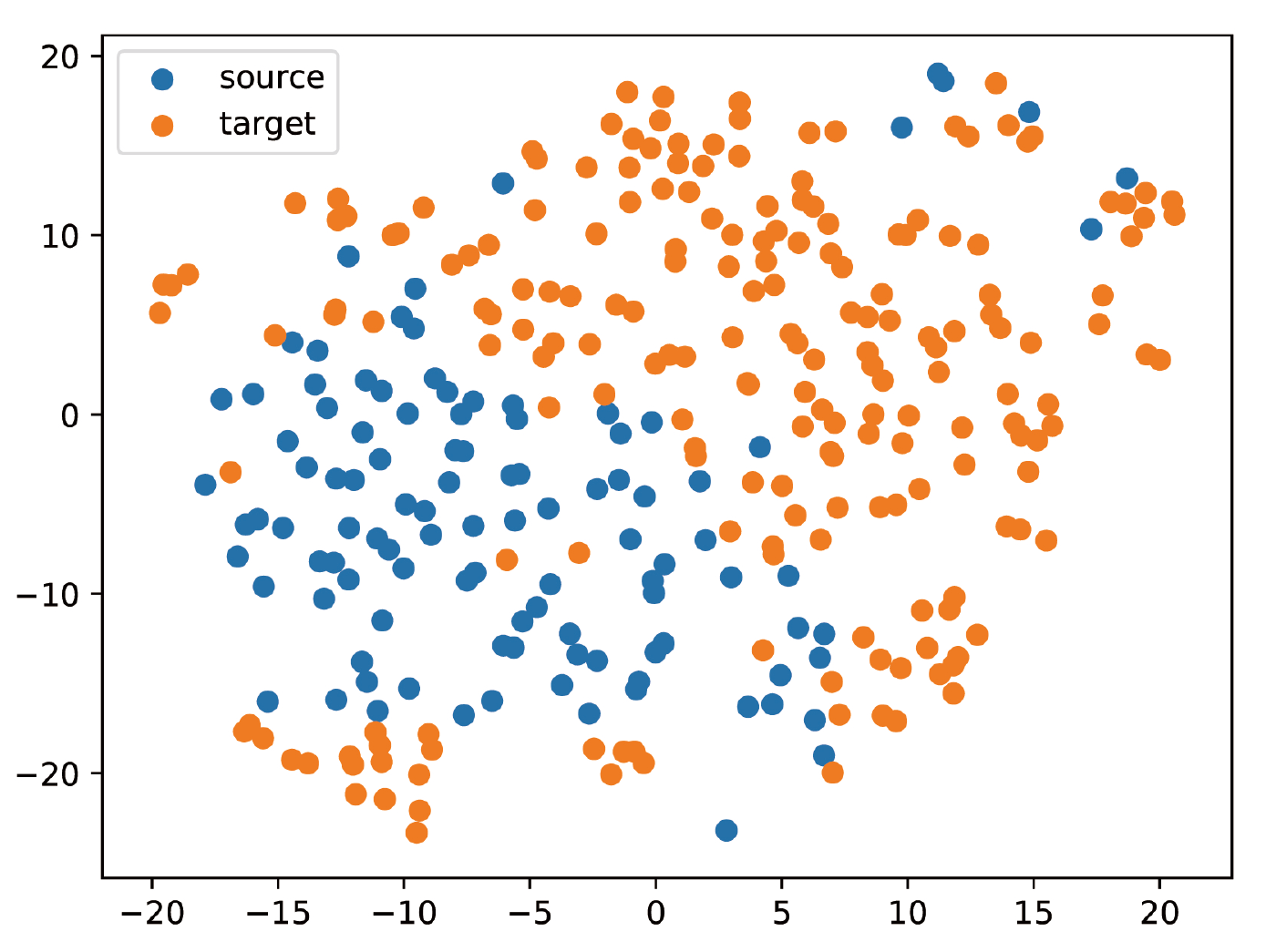}
		}
		\subfigure[Class distributions]{
			\label{fig:tsneR}
			\includegraphics[width=0.47\linewidth]{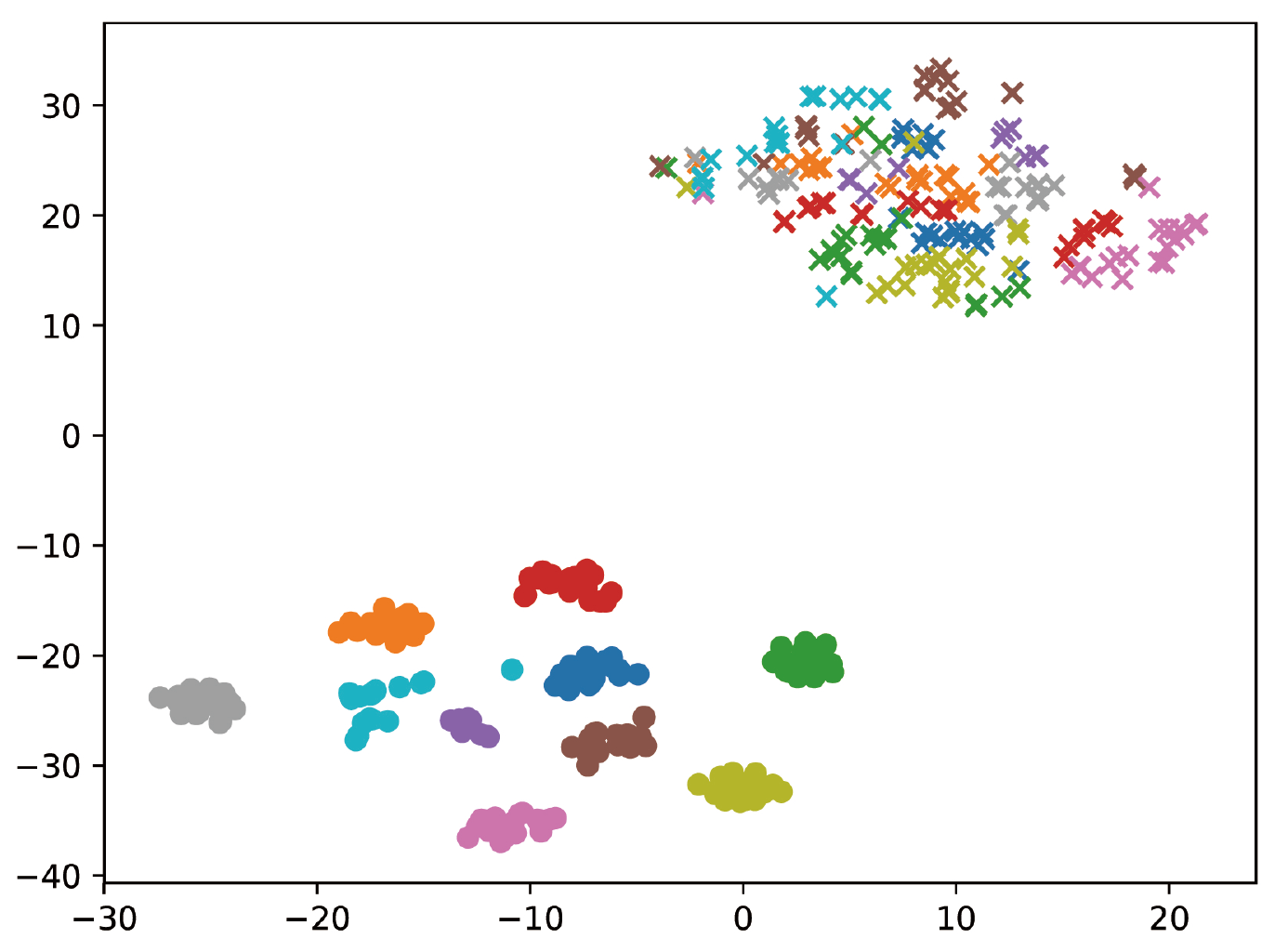}
		}
		\caption{The distributions of disentangled representations \emph{w.r.t.} (a) domains and (b) classes in the test set. Marker $\bullet$ and $\times$ denote task-relevant and task-irrelevant representations, respectively. The colors in Fig.~\ref{fig:tsneR} denote different classes.}
		\vspace{-1em}
	\end{figure}

	\textbf{Visualization.}
	We visualize the distributions of the learned task-relevant and task-irrelevant representations by t-SNE. In Fig.~\ref{fig:tsneL}, task-relevant representations from the source and target domains are clearly separated, which demonstrates that our approach 
	does not alter the distribution of data, but divides it into two parts from the perspective of the task. \blue{Fig.~\ref{fig:tsneR} shows the distributions of task-relevant and task-irrelevant representations in target domain, 
		which further proves the correctness of our disentanglement based on the following observations. (1) The two disentangled parts are clearly separated. (2) The task-relevant representations are clustered according to their predicted labels. (3) The task-irrelevant representations are not clearly separable according to their predicted labels.}
	
	\subsection{Quantitative Results of Domain Adaptation}
	We compare DTDN to existing domain adaptation methods in retrieval tasks on Person-reID and Vehicle-reID, and classification tasks on Office-31 and digits datasets.
	
	\textbf{Person-reID. } The results of Person-reID are shown in Tab.~\ref{table:person-reid}.
	%
	\red{For Duke $\to$ Market adaptation, Rank-1 accuracy achieves 75.2\% and \emph{m}AP is 43.9\%. These two results are then improved to 80.1\% and 47.1\% respectively when IN is applied. As for Market $\to$ Duke, DTDN outperforms existing methods with the Rank-1 and \emph{m}AP of 65.4\% and 44.9\%, respectively. Applying IN further improves these two results to 70.0\% and 47.7\%.} Note that the above improvements are achieved with neither generating extra images~\cite{wei2018person,deng2018image} nor exploiting extra attribute information~\cite{wang2018transferable}.

	\textbf{Vehicle-reID.}
	The results of Vehicle-reID are shown in Tab.~\ref{table:vehicle-reid}. Since we are the first to report \emph{unsupervised} domain adaptation results for Vehicle-reID tasks, the proposed model is compared to \emph{supervised} methods. For VehicleID $\to$ VeRi adaptation, \red{we achieve the Rank-1 accuracy of 74.5\%, which is then improved to 75.3\% by applying IN. As for VehicleID, we achieves 52.2\%, 49.6\%, 44.3\% on Rank-1 in the small, medium, and large test set, respectively.} And IN improves the results to 52.8\%, 50.7\%, and 45.7\%.
	
	\textbf{Office-31.} %
	To demonstrate general applicability of our approach, experiments on the Office-31 dataset in an open-set setting are conducted, where the target domain has an additional class that is unknown in the source domain. Results are shown in Tab.~\ref{table:office31}, where OS and OS* denotes accuracy averaged over all and known classes, respectively. Results of competing approaches are from previous work~\cite{liu2019separate}.
	
	Note that DTDN is mainly designed for retrieval tasks but not classification ones~\cite{conf/iccv/BustoG17,DBLP:conf/eccv/SaitoYUH18,liu2019separate}.
	We therefore detect the unknown class by comparing the maximum probability of belonging to known classes to a threshold (0.3 is used for all experiments). Experiments show that our method still perform well in classification tasks. To explain, our high OS* accuracy shows that our approach can effectively identify unknown classes, and the comparable OS accuracy shows that the target domain's unknown class and the source domain's known classes are correctively separated.
	
	\textbf{Digits. }%
	Tab.~\ref{table:digits} reports results on three digits datasets.
	We achieve the Rank-1 of 99.08\% for M$\to$U, 99.15\% for U$\to$M, and 98.15\% for S$\to$M, which outperform the baseline by 3\%--14\% and are only 0.14\%--1.64\% lower than supervised approaches. It shows that our model can also perform well in close-set settings. Note that, due to the different backbone networks used by the different methods, we show the results of \emph{Supervised Learning} (backbone \emph{ResNet-20} trained and tested in the labeled target domain) and \emph{Source Only} (backbone network trained in the labeled source domain and tested in the unlabeled target domain) of our backbone network for comparison.
	\subsection{Ablation Study and Discussions}
	The ablation study in Tab.~\ref{table:ablation} investigates the effectiveness of each component by disabling it in the Duke $\to$ Market adaptation task.
	%
	Firstly, the accuracy drops significantly when $L_{task}$ is disabled, which means DTDN cannot properly extract features and finish the task, as the guidance of a labeled source domain is missing.
	%
	Secondly, even without $L_{neighbor}$, we observe that DTDN achieves results better than the baseline. This is due to the fact that $L_{task}$ and $L_{agree}$ can reject task-irrelevant information yet fail to guarantee the domain-sharing and transferability of the remaining task-relevant components.
	%
	Thirdly, for $L_{agree}$, we recognize its additional improvements over the rejection of task-irrlevant information in the target domain, and highlight the observation that $L_{task}$ and $L_{neighbor}$ (\emph{i.e.}, w/o ${L_{agree}})$ can achieve suboptimal accuracy by learning task-relevant information that is both useful for the task and transferable across domains.
	%
	\begin{table}[tp]
		\caption{Ablation Study}
		\begin{adjustbox}{max width=1.0\linewidth}
			\begin{tabular}{c|c|c|c|c|c}
				\hline
				\multirow{1}{*}{Source $\to$ Target} & \multicolumn{5}{c}{DukeMTMC-reID $\to$ Market-1501} \\
				\hline\hline
				Metric (\%) & Rank-1  & Rank-5 & Rank-10 &Rank-20 &\emph{m}AP     \\ \hline
				Baseline (source only)     & 46.6 & 63.9 & 70.1 & 76.2 & 18.6     \\
				DTDN w/o $L_{task}$ & 2.0 & 5.0 & 7.5 & 11.4 & 0.5  \\
				DTDN w/o $L_{neighbor}$ & 59.2 & 76.6 & 82.2 & 86.8 & 29.2     \\
				DTDN w/o $L_{agree}$ & 72.3 & 83.9 & 87.3 & 90.4 & 41.5     \\ \hline
				DTDN & 75.2     & 85.4    & 88.9     & 92.0     & 43.9      \\
				DTDN+IN (Ours) & \textbf{80.1}     & \textbf{87.8}    & \textbf{90.8}     & \textbf{93.6}     & \textbf{47.1}      \\
				\hline
			\end{tabular}
		\end{adjustbox}
		\label{table:ablation}
		\vspace{-0.5em}
	\end{table}
	\begin{table}[tp]
		\caption{Top-1 Accurancy (\%) on Person-reID}
		\begin{adjustbox}{max width=1.0\linewidth}
			\begin{tabular}{c|c|c|c|c|c}
				\hline
				TR:TI & 128:384 & 256:256 & 320:192 & 512:0 & Dynamic \\
				\hline
				Accuracy & 39.7 & 54.6 & 63.2 & 58.3 & \textbf{65.4}      \\
				\hline
			\end{tabular}
		\end{adjustbox}
		\label{table:Dynamic}
		\vspace{-2em}
	\end{table}
	
	\blue{To show that our dynamic network eliminates the cumbersome settings of complex hyper-parameters caused by the variable ratio of task-relevant (TR) and task-irrelevant (TI) representations across tasks, the empirical hardness of finding a well-performed ratio (TR:TI) by hand is shown in Tab.~\ref{table:Dynamic}, which report several attempts in the Market$\to$Duke Person-reID task.}
	%
	%
	As shown in the table, performance varies among different ratios and the optimal ratio for one particular task is usually no longer optimal for another task, nonetheless, the proposed dynamic network can well adaptively learn the optimal ratio for different tasks all by itself.

	\section{Conclusion}
	In this paper, we propose a dynamic task-oriented disentangling network (DTDN) to learn disentangled representations in an end-to-end fashion for UDA, which implicitly avoids the use of generative models or decoders.
	We break the concept of task-orientation into task-relevance and task-irrelevance, and disentangle the representations from different domains into task-relevant and task-irrelevant components, which guarantees the learned domain-transferable information is also class-discriminative. 
	This is achieved by shuffling and swapping the task-irrelevant components of both domains. These two components are regularized by a group of task-specific objective functions across domains.
	It also implicitly preserves complete and meaningful information without generative models or decoders, which massively reduces the training cost.
	Experiments of digits recognition, Office-31 classification, and person/vehicle reID tasks demonstrate DTDN's superior performance in both close-set and complicated, open-set scenarios.


	\bibliographystyle{ieee_fullname}
	\bibliography{dual}

\end{document}